\documentclass[11pt,a4paper]{article}
\usepackage[hyperref]{emnlp2020}
\usepackage{times}
\usepackage{latexsym}
\usepackage{graphicx}

\usepackage{microtype}

\aclfinalcopy 

\title{LynyrdSkynyrd at WNUT-2020 Task 2: Semi-Supervised Learning for Identification of Informative COVID-19 English Tweets}

\author{Abhilasha Sancheti$^*$\\
  University of Maryland\\
  College Park, USA\\
  \texttt{sancheti@cs.umd.edu} \\\And
  Kushal Chawla\thanks{\hspace{0.2cm}Denotes equal contribution.} \\
  University of Southern California \\
  Los Angeles, USA \\
  \texttt{kchawla@usc.edu}\\ \And
  Gaurav Verma\\
  Adobe Research\\
  Bangalore, India\\
  \texttt{gaverma@adobe.com}
  }
\date{}

\begin{document}
\maketitle
\begin{abstract}
We describe our system for WNUT-$2020$ shared task on the identification of informative COVID-19 English tweets. Our system is an ensemble of various machine learning methods, leveraging both traditional feature-based classifiers as well as recent advances in pre-trained language models that help in capturing the syntactic, semantic, and contextual features from the tweets. We further employ pseudo-labelling to incorporate the unlabelled Twitter data released on the pandemic. Our best performing model achieves an F1-score of $0.9179$ on the provided validation set and $0.8805$ on the blind test-set.
\end{abstract}
\section{Introduction}

The 2020 edition of Workshop on Noisy User-generated Text (W-NUT 2020) hosted a shared task on `Identification of Informative COVID-19 English Tweets'. The task involves automatically identifying whether an English Tweet related to the novel coronavirus (COVID-19) is `informative' or not. For a tweet to be considered informative in this context, it should provide information about recovered, suspected, confirmed, and death cases as well as location or travel history of the cases. The goal for developing such an automated system is to help track the development of the COVID-19 outbreak and to provide users the information related to the virus, e.g. any new suspicious/confirmed cases near/in the users' regions.

Aligned with the goals of this shared task, our paper details the use of state-of-the-art natural language processing techniques for this task.
We experiment with a variety of methods, ranging from feature-based classifiers to leveraging recent advances in pre-trained neural architectures (Section \ref{sec: system}). To further improve performance, we incorporate unlabelled tweets released on COVID-19 via masked language modelling and pseudo-labelling techniques. Our best performing model is an ensemble that uses Logistic Regression to combine the output probabilities of several base classifiers (Section \ref{sec: results-analysis}). We further analyze the impact of pre-processing and semi-supervision through ablation studies. Through our qualitative adversarial analysis, we show how the predictions of BERT model are sensitive towards specific tokens such as `confirmed case' or even locations and numerals, which also guides our data pre-processing steps.

\section{Data Preprocessing} \label{sec: dataset}

We conduct classification experiments on the COVID-19 Tweets dataset \cite{covid19tweet} provided for the shared task. The data split consists of $7000$ tweets for training, $1000$ in the validation set and $12000$ in the blind test-set.

We start by lowercasing all the tweets, and replacing all urls with `httpurl' and all usernames with `@user'. We then normalize all characters and pictograms.  Additionally, we remove bad symbols (e.g. from a different language), any html tags, duplicated symbols or characters (like dots, question marks, special symbols, dashes or exclamations), and unnecessary underscores in the tokens. We also isolate punctuations, expand contractions (e.g. there'll to there will), and replace leet alphabets with their correct English versions using leet vocabulary\footnote{\url{https://en.wikipedia.org/wiki/Leet}}. We convert emojis into their text form and finally normalize the different ways in which `COVID-19' is written (e.g., covid-19, covid2019, covid19, or covid 2019) by replacing each occurrence with `covid19'. These steps help to limit the vocabulary for better learning.

On top of the above steps, we create different versions of the dataset as described below.

\noindent\textbf{Cleaned:} No further processing is done.

\noindent\textbf{NUM-replaced:} All the numerals in the dataset (except 19 in `covid19') are replaced by NUM followed by the number of digits in the number. 

\noindent\textbf{LOC-replaced:} We use the Python Spacy library\footnote{Spacy's en\_core\_web\_sm module.} to get the named entity tags for each token in a tweet and replace the token having a `GPE' (geopolitical entities i.e. countries, states, or cities) tag with `LOC', so that the informativeness of a tweet is agnostic of a particular location.\\
\textbf{NUM-LOC-replaced:} We replace both numerals and tokens with `GPE' tags in this version.

The motivation behind different versions of the dataset comes from our analysis (presented in Section \ref{sec: results-analysis}) of the model trained on the \textbf{Cleaned} version of the dataset. We show how the model predictions are sensitive towards locations and numerals in a tweet, as a result of the patterns in the dataset. Hence, using NUM and LOC helps mitigate these issues. Following this pre-processing, we train a variety of machine learning models to classify tweets into two classes: informative and uninformative. We use the data split provided by the task organizers throughout to ensure consistency.

\section{System Description} \label{sec: system}

The system submitted for the shared-task is an ensemble of a variety of machine learning models which we discuss below in detail.

\subsection{Fine-tuning pre-trained models}
\label{sec: bert-models}
Motivated by their state-of-the-art performance on several downstream natural language processing tasks, we use pre-trained models like BERT ~\cite{devlin2018bert} for our classification task. BERT models learn a contextual representation of an input text which helps in representing the semantic information contained in an input. Borrowing notations from \citet{sun2019fine}, we leverage the pre-trained model in the following ways.

\noindent\textbf{BERT-FiT:} Following the standard fine-tuning pipeline~\cite{devlin2018bert}, the model is initialized with a pre-trained architecture and is further \textbf{Fi}ne-\textbf{T}uned on the provided labelled dataset.

\noindent\textbf{BERT-ITPT-FiT:} Inspired by the improvements using with\textbf{I}n \textbf{T}ask \textbf{P}re-\textbf{T}raining~\cite{sun2019fine}, we further train the out-of-the-box pre-trained model on our training dataset using the masked language modeling (MLM) objective for $3$ epochs. Thereafter, the model is fine-tuned for classification using the associated ground-truth labels.

\noindent\textbf{BERT-IDPT-FiT:} Extending the above approach, 
we extract a total of $27,388$ tweets related to the pandemic using the Twitter API\footnote{\url{https://developer.twitter.com/en/docs/labs/tweets-and-users/quick-start/get-tweets}} for \textbf{I}n-\textbf{D}omain \textbf{P}re-\textbf{T}raining.  We use the tweet IDs released with the Covid-19 tweets dataset~\cite{781w-ef42-20} and the WNUT shared task on Event Extraction\footnote{\url{http://noisy-text.github.io/2020/extract_covid19_event-shared_task.html}}.
The extracted tweets are pre-processed in the same manner as described in Section \ref{sec: dataset} and augmented with the training set to further train the pre-trained model on the MLM loss for $5$ epochs. Once trained, we then fine-tune the model on the classification task using the provided labelled training set. As we show later (Table \ref{tab: evaluation}), leveraging additional in-domain tweets further helps in improving the performance on our task.

We primarily use `bert-base-uncased' architecture for all our experiments and train separate models\footnote{We use huggingface library available at \url{https://github.com/huggingface/transformers} for fine-tuning of pre-trained models. } for each version of the dataset as well as each variant of the BERT model. All the models are fine-tuned for $20$ epochs with the maximum input length set at $100$ and dropout at $0.3$.

We also experimented with other pre-trained models: BERT-Large~\cite{devlin2018bert}, RoBERTa~\cite{liu2019roberta}, XLNet~\cite{yang2019xlnet}, and Covid-BERT~\cite{muller2020covid} but did not find any visible difference in performance. Hence, we restricted our experiments to BERT-base owing to its smaller size and consequently, lesser consumption of computational resources.

\subsection{Classification using fastText (fastText)}
\citet{joulin2017bag} proposed fastText – a simple and efficient baseline for text classification that uses bag of n-gram features to capture partial information about the local word order. We use the fastText library to train a classifier for our task. We train the model for $10$ epochs with a learning rate initialized at $0.1$ and the maximum length of n-grams is set to $3$. 
\subsection{Most-frequent N-gram features (ngram)}
We also build our own implementation of n-gram features. Using the \textbf{NUM-LOC-replaced} version of the dataset, we extract the most frequent $5,000$ unigrams, bigrams, and trigrams for each class and use the presence or absence of these n-grams in a given tweet, as features. These features are used as input to a feed-forward neural network, with two hidden layers of size $64$, dropout of $0.1$, and activation as ReLU. Although the feature vector is highly sparse, we find this approach to perform reasonably well on our task. 
\subsection{Leveraging the Universal Sentence Encoder (USE)}
We leverage the Universal Sentence Encoder~\cite{cer2018universal} to get sentence embeddings of the tweets. These $512$-dimensional embeddings are used as input features to a feed-forward neural network identical to the one described above for n-gram features. Sentence embedding based classifiers also take into account the contextual (short-range as well as long-range) information of each word as well as ordering of the words, unlike the n-gram based models, performing well in practice. 

\subsection{Hand-crafted features (HCF)}
Next, leveraging the advances in affective computing, we build a total of $150$ hand-crafted features, comprising both syntactic and affect features. Syntactic features include statistics from the tweet based on the counts of punctuation marks along with NER and POS tags. We also include text readability metrics from the textstat toolkit\footnote{\url{https://pypi.org/project/textstat/}}. On the other hand, affect features are based on existing lexicons in the literature. Specifically, we use Warriner VAD wordlists~\cite{warriner2013norms}, formality word-lists~\cite{brooke2010automatic}, PERMA model~\cite{seligman2012flourish}, Temporal word lists~\cite{park2015living}, EmoLex~\cite{mohammad2013nrc}, and lastly, LIWC~\cite{pennebaker2001linguistic}. All the features are concatenated to form a feature vector for each tweet which is used as an input to the feed-forward network described above.
\subsection{Employing Pseudo-Labelling (PL)}
In order to leverage the unlabelled data for feature-based methods, we resort to pseudo-labelling. The models are trained for at most $50$ epochs in the following manner: After every $10$ epochs, we select the checkpoint with the best F1-score saved till now and use it to predict labels on the collected unlabelled tweets (same as described in Section \ref{sec: bert-models}). We randomly pick $1000$ tweets on which the prediction confidence is above $0.99$ and consider the predicted labels for these tweets as pseudo ground-truth labels. Hence, these $1000$ tweets are removed from the set of unlabelled data and instead added to the training dataset for future epochs. The training continues from the best checkpoint found till now.

In Section \ref{sec: results-analysis}, we show improvements due to the use of pseudo-labelling across all the methods based on fixed feature vectors. We also tried pseudo labelling with BERT. However, using MLM to incorporate unlabelled tweets (\textbf{BERT-IDPT-FiT}), performs better than pseudo-labelling in that case.
\subsection{Ensemble Model}
Our final model is an ensemble of the base classifiers described above. To build an ensemble, we train a logistic regression model using the output probabilities from a subset of the base classifiers, as features. Further, we tune the threshold above which the model would output the class as `Informative', and otherwise output `Uninformative'. The ensemble model without any threshold tuning is referred to as \textbf{Ensemble}. The model with threshold tuning is referred to as \textbf{Ensemble-TT}, that uses a threshold value of $0.5168$ instead of $0.5$.

\begin{table}[t]
\centering
\scalebox{0.7}{
\begin{tabular}{c|c|ccc}
\hline
{\textbf{Model}} & {\textbf{Accuracy$\uparrow$}} & {\textbf{Precision$\uparrow$}} & {\textbf{Recall$\uparrow$}} & {\textbf{F1-score$\uparrow$}}\\
\hline
\textbf{Majority} & $0.5280$ & $0.0000$ & $0.0000$ & $0.0000$ \\
\hline
\textbf{HCF-PL} & $0.7460$ & $0.7410$ & $0.7104$ & $0.7252$ \\
\textbf{USE-PL} & $0.7756$ & $0.7196$ & $0.8601$ & $0.7834$\\
\textbf{fastText} &$0.8083$  &$0.8372$  & $0.7373$ & $0.7840$\\
\textbf{ngram-PL} & $0.8090$ & $0.7932$ & $0.8072$ &$0.7997$ \\
\textbf{BERT-FiT} & $0.9010$ &$0.8768$  &$0.9201$  &$0.8977$  \\ 
\textbf{BERT-ITPT-FiT} & $0.9010$ & $0.8716$& $0.9272$ &  $0.8984$\\
\textbf{BERT-IDPT-FiT} & $0.9070$ &$0.8736$ & $\mathbf{0.9392}$ &$0.9050$  \\
\hline
\textbf{Ensemble} & $\mathbf{0.9209}$ & $\mathbf{0.9034}$ & $0.9322$ & $0.9176$ \\
\textbf{Ensemble-TT} & $\mathbf{0.9209}$ & $0.9002$ &$0.9364$ &$\mathbf{0.9179}$  \\ 
\hline
\end{tabular}}
\caption{Performance of various models on Precision, Recall and F1-score for Informative class and model accuracy on the validation dataset. The numbers are averaged over $3$ runs. BERT-based and fastText models were trained on different pre-processed versions of the dataset. We report the scores for the corresponding best performing configuration above.}
\label{tab: evaluation}
\end{table}

\section{Results and Analysis}
\label{sec: results-analysis}
\begin{figure}[th]
    \centering
    \includegraphics[width=0.45\textwidth,height=4.5cm]{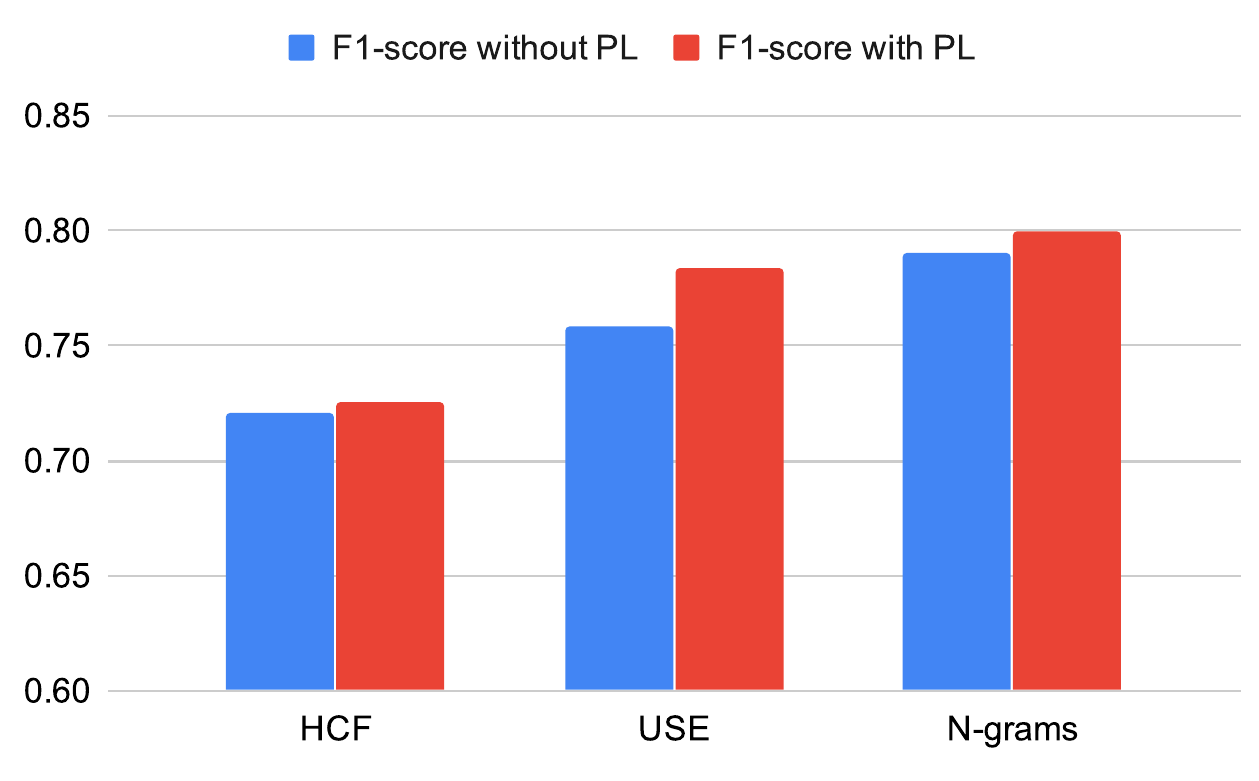}
    \caption{Effect of pseudo-labelling on F1-score for the Informative class. Pseudo-labelling improves the performance for all three methods.}
    \label{fig:pseudo}
\end{figure}
\begin{figure}[th]
    \centering
    \includegraphics[width=0.45\textwidth,height=4.8cm]{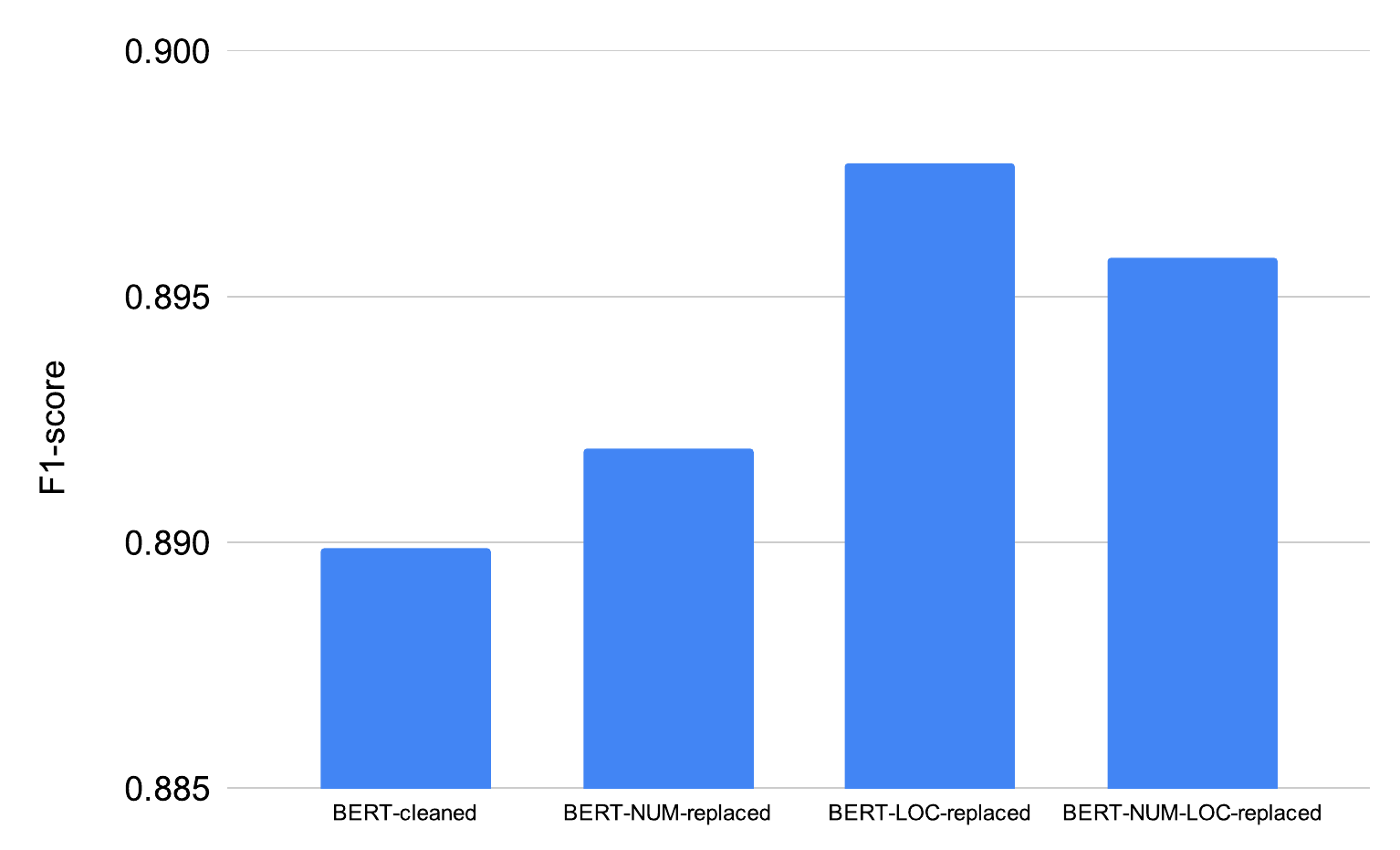}
    \caption{Impact of different ways of data pre-processing on F1-score using the \textbf{BERT-FiT} model.}
    \label{fig:pre-processing}
\end{figure}
The primary evaluation metric for the shared task is F1-score for the `Informative' class. The other metrics used are the Precision and Recall for Informative class and the overall accuracy of the model.

We first illustrate the benefit of employing pseudo-labelling in Figure \ref{fig:pseudo}. All three methods, namely \textbf{HCF}, \textbf{USE} and \textbf{ngrams} achieve better F1-scores by leveraging the unlabelled data, along with the provided training dataset. Hence, we next compare these models with other baselines and BERT-based pre-trained methods in Table \ref{tab: evaluation}. 
All models easily beat the majority baseline, attesting that the models are learning useful patterns from the data. While feature-based methods perform reasonably well with \textbf{ngram-PL} achieving an F1-score close to $0.8$, they are outperformed by BERT-based methods with approximately a $10\%$ gain. High performance of \textbf{BERT-IDPT-FiT} shows that incorporating additional in-domain data using the MLM objective improves the performance. Finally, an ensemble using logistic regression achieves the best precision, resulting from a reduction in the number of false positives, while still maintaining the same recall. Our best performing model, \textbf{Ensemble-TT} achieves an F1-score of $0.9179$ on the validation dataset and $0.8805$ on the blind test-set.
\begin{table}[th]
\centering
\scalebox{0.7}{
\begin{tabular}{p{7.7cm}|p{1.5cm}}
\hline
{\textbf{Input:} the taslee palm city estate in maitama , abuja , has alerted ... } & {\textbf{Prediction}} \\
\hline
... its residents. & $0$\\
\hline
... its residents of a case .
&$1$\\
\hline
... its residents of a confirmed case .
&$1$\\
\hline
... its residents of coronavirus.
&$0$\\
\hline
... its residents of a case of coronavirus .
&$1$\\
\hline
... its residents of a confirmed case of coronavirus .&$1$\\
\hline
... its residents of malaria.
&$0$\\
\hline
... its residents of a case of malaria .
&$1$\\
\hline
... its residents of a confirmed case of malaria .
&$1$\\ \hline
\end{tabular}}
\caption{\textbf{BERT-FiT} predictions on artificially-created examples, inspired from an instance in the validation dataset. $0$ $(1)$ refers to Uninformative (Informative) class.}
\label{tab: adversarial}
\end{table}
\begin{table}[th]
\centering
\scalebox{0.7}{
\begin{tabular}{p{6.5cm}|p{1.5cm}}
\hline
{\textbf{Input}} & {\textbf{Prediction}} \\
\hline
his family members got infected in santa clara .
&$1$\\
\hline
his family members got infected in rohini .
&$0$\\
\hline
5 family members got infected in rohini .
&$1$\\
\hline
5 family members are healthy in rohini .
&$1$\\ \hline
\end{tabular}}
\caption{\textbf{BERT-FiT} predictions on artificially-created examples, depicting bias towards numerals and locations. $0$ $(1)$ refers to Uninformative (Informative) class.}
\label{tab: num-loc-bias} 
\end{table}

\noindent \textbf{Qualitative Adversarial Analysis:} In Table \ref{tab: adversarial}, we show the sensitivity of \textbf{BERT-FiT} model towards specific tokens such as `confirmed' and `case' in the tweets. Using these tokens governs the output predictions, regardless of whether the tweet is talking about covid-19 or an arbitrarily chosen disease, malaria. This is justified since the dataset mostly contains the tweets related to the pandemic, but suggests to exercise caution while using such models for downstream monitoring applications. Further, in Table \ref{tab: num-loc-bias}, we show sensitivity towards numeric and location tokens in the tweets. Mere change of location to a less frequent one in the dataset or use of numerals inverts the model predictions, regardless of whether the tweet is actually informative or not. This observation infact inspires our data-pre-processing stages, where we mask the numeric and location tokens from the dataset. We investigate this further in Figure \ref{fig:pre-processing}, which establishes that effective pre-processing can help mitigate these biases, while keeping the performance at-par or better on our classification task.


\section{Conclusion} \label{sec: conclusion}
In this paper, we describe our system to identify informative COVID-19 English tweets. We find that an ensemble model which uses a logistic regression to combine the predictions of a variety of feature-based to neural methods achieves the best performance on the shared task. Our analysis shows that incorporating unlabelled tweets results in consistent performance gains. We show how the trained model can be sensitive to specific tokens in the tweets, and hence, advice for exercising caution while deploying machine learning models for downstream monitoring applications.

\bibliographystyle{acl_natbib}
\bibliography{anthology,emnlp2020}



\end{document}